\documentclass[11pt,a4paper]{article}
\usepackage[hyperref]{emnlp2020}
\usepackage{times}
\usepackage{latexsym}

\usepackage{microtype}
\usepackage{times}
\usepackage{xcolor}
\usepackage{natbib}
\usepackage{soul}
\usepackage{tikz}
\usepackage{tikz-dependency}
\usetikzlibrary{positioning}
\usepackage[utf8]{inputenc}

\usetikzlibrary{calc,intersections}
\usepackage[nocenter]{qtree}

\usepackage{CJKutf8}
\usepackage{url}
\usepackage{tipa}

\usepackage{amsmath,amsthm,amssymb,amsfonts,bbm,stmaryrd}
\usepackage{enumitem}

\usepackage{textcomp}
\usepackage{multirow} 
\usepackage{graphicx} 
\usepackage{subcaption}
\usepackage{booktabs} 
\usepackage[normalem]{ulem}
\usepackage{makecell}
\usepackage{courier}
\usepackage{bbm}
\usepackage{algorithm,algpseudocode}
\usepackage{adjustbox}
\usepackage{tabularx}
\usepackage{booktabs}
\usepackage{setspace}
\usepackage{todonotes}
\usepackage{threeparttable}

\usepackage{listings}
\usepackage{pifont}
\usepackage{color}
\usepackage{bm}
\usepackage{ulem}
\usepackage{array}

\DeclareMathOperator*{\argmax}{argmax}

\usepackage{pifont}
\newcommand{\xmark}{\ding{55}}%

\newcommand{\mb}{\mathbf}

\newcommand{\y}{\mathbf{y}}
\newcommand{\z}{\mathbf{z}}

\newcommand{\bz}{\mathbf{z}}

\newcommand{\argument}[1]{\textcolor{orange}{#1}}

\newcommand{\role}[1]{\textcolor{red}{#1}}

\usepackage{siunitx}
\usepackage{ragged2e}
\newcolumntype{R}[1]{>{\RaggedLeft\arraybackslash}p{#1}}

\newcolumntype{P}[1]{>{\RaggedRight\arraybackslash}p{#1}}

\newcommand*{\MinNumbera}{0.0}%
\newcommand*{\MaxNumbera}{1.0}%

\newcommand{\Cpp}[2]{\setlength{\fboxsep}{2pt}\pgfmathsetmacro{\PercentColora}{65.0*(#2-\MinNumbera)/(\MaxNumbera-\MinNumbera)}\colorbox{green!\PercentColora!white}{\strut #1}}

\newcommand*{\MinNum}{0.0}%
\newcommand*{\MaxNum}{.82}%
\newcommand{\MRpp}[2]{\setlength{\fboxsep}{2pt}\pgfmathsetmacro{\PercentColora}{65.0*(#2-\MinNum)/(\MaxNum-\MinNum)}\colorbox{red!\PercentColora!white}{\strut #1}}
\newcommand{\MBpp}[2]{\setlength{\fboxsep}{2pt}\pgfmathsetmacro{\PercentColora}{65.0*(#2-\MinNum)/(\MaxNum-\MinNum)}\colorbox{blue!\PercentColora!white}{\strut #1}}

\tikzset{block/.style={
		rounded corners, fill=blue!15,
		anchor=north west}}
\tikzset{line/.style={
		-latex,
		line width=.5mm,
		black!65}}
\tikzset{annotation/.style={
		align=left,
		anchor=north west,
		draw=black!75,
		minimum height=10mm,
		minimum width=10mm}}
\tikzset{hmodule/.style={
		rounded corners,
		align=center,
		anchor=north west,
		font=\Large,
		fill=blue!20,
		draw=lightgray!75,
		minimum height=10mm,
		minimum width=20mm}}
\tikzset{vmodule/.style={
		rounded corners,
		align=center,
		rotate=90,
		anchor=north west,
		font=\Large,
		fill=blue!20,
		draw=lightgray!75,
		minimum height=10mm,
		minimum width=20mm}}
\tikzset{hvector/.style={
		rounded corners=2.5mm,
		align=center,
		anchor=north west,
		fill=gray!5,
		draw=black!75,
		minimum height=5mm,
		minimum width=20mm}}
\tikzset{vvector/.style={
		rounded corners=2.5mm,
		align=center,
		rotate=90,
		anchor=north west,
		fill=gray!5,
		draw=black!75,
		minimum height=5mm,
		minimum width=20mm}}

\aclfinalcopy 

\title{Unsupervised Transfer of Semantic Role Models \\ from  Verbal to Nominal Domain}

\newcommand{\uoa}{\normalfont \text{\textipa{\ae}}}
\newcommand{\uoe}{\normalfont \text{\textipa{E}}}

\author{Yanpeng Zhao$^{\uoe}$ \\
	$^{\uoe}$ILCC, University of Edinburgh \\
	\tt{yanp.zhao@ed.ac.uk} \\\And
	Ivan Titov$^{\uoe\uoa}$\\
	$^{\uoa}$ILLC, University of Amsterdam \\
	\tt{ititov@inf.ed.ac.uk} \\}

\date{}

\begin{document}
\maketitle
\begin{abstract}
Semantic role labeling (SRL) is an NLP task involving 
the assignment of predicate arguments to types, called {\it semantic roles}.
Though research on SRL has primarily focused on verbal predicates and many resources available for SRL provide annotations only for verbs, semantic relations are often triggered by other linguistic constructions, e.g., nominalizations. In this work, we investigate a transfer scenario where we assume role-annotated data for the source verbal domain but only unlabeled data for the target nominal domain. Our key assumption, enabling the transfer between the two domains, is 
that {\it selectional preferences} of a role (i.e.,
preferences or constraints on the admissible arguments) do not strongly depend on whether the relation is triggered by a verb or a noun.
For example, the same set of arguments can fill the \textit{Acquirer} role
for the verbal predicate `acquire' and its nominal form `acquisition'.
We approach the transfer task from the variational autoencoding  perspective.
The labeler serves as an encoder (predicting role labels given a sentence),
whereas selectional preferences are captured in the decoder component (generating arguments for the predicting roles). Nominal roles are not labeled in the training data, and the learning objective instead 
pushes the labeler to assign roles predictive of the arguments. Sharing the decoder parameters across the domains 
encourages consistency between labels predicted for both domains and facilitates the transfer.
The method substantially outperforms baselines, such as unsupervised  and `direct transfer' methods, on the English CoNLL-2009 dataset.\footnote{Our code is available at \href{https://github.com/zhaoyanpeng/srltransfer}{https://git.io/JfO9q}.}
\end{abstract}

\section{Introduction}

Semantic role labeling~\citep{gildea2002automatic} methods detect the underlying predicate-argument structures of sentences, or, more formally, assign {\it semantic roles} to
arguments of predicates:\footnote{In this work, we consider the popular {\it dependency-based} SRL version~\cite{hajivc2009conll},
i.e., only syntactic heads of argument phrases are marked with roles rather than entire argument spans (e.g., {\it `book'} rather than {\it `the book'}).}

\begin{center}\label{sent:rabbit}
	
	$[\text{Pinocchio}]_{\text{A0}}$ \textit{traded} the $[\text{book}]_{\text{A1}}$ for a $[\text{ticket}]_{\text{A3}}$.
\end{center}
In this sentence,
\textit{Pinocchio} is labeled as {\text A0}, indicating that it is an agent (`a trader') of the \textit{trading} event, whereas
\textit{book} is marked as a patient ({\text A1}, `an entity being traded').
Semantic-role structures have been shown effective in many NLP tasks, including machine translation~\citep{marcheggiani2018exploiting},
question answering~\citep{shen2007using},
and summarization~\citep{khan2015framework}.

\begin{figure}
\centering
\resizebox{.925\linewidth}{!}{%
\begin{tikzpicture}[node distance=0mm]
\node[block]  (V) at (1,10) {\begin{minipage}{8cm}
	\textbf{Labeled Verbal Data}\newline 
	He added that until [\argument{now}]$_{\text{\role{TMP}}}$ the [\argument{Japanese}]$_{\text{\role{A0}}}$ have [\argument{only}]$_{\text{\role{ADV}}}$ \textbf{\textit{acquired}} equity [positions]$_\text{\role{A1}}$ 
	in U.S. biotechnology companies .
	\end{minipage}};

\node[rounded corners, fill=brown!5,draw=lightgray!75,anchor=north west,minimum width=1.75cm,minimum height=1.5cm] (M) at (1.0, 7.5) {\textbf{MODEL}};

\node[rounded corners, fill=white!10,draw=lightgray!75,anchor=north west,minimum width=2cm,minimum height=1.5cm,align=left] (O) at (3.3, 7.5) {
	[\argument{Sony}]$_{\text{\role{A0}}}$,\hspace{.0em} [\argument{recent}]$_{\text{\role{ADV}}}$ ,\hspace{.0em} [\argument{Pictures}]$_{\text{\role{A1}}}$
};

\node[block] (N) at (1, 5.5) {\begin{minipage}{8cm}
	\textbf{Unlabeled Nominal Data}\newline 
	After [\argument{Sony}] Corp. ’s [\argument{recent}] headline - grabbing \textbf{\textit{acquisition}} of Columbia [\argument{Pictures}] , many say it makes good political sense to lie low .
	\end{minipage}};

\draw [-latex,line width=.3mm,black!65] (V) edge (M);
\draw [-latex,line width=.3mm,black!65] (M) edge (O);
\draw [-latex,line width=.3mm,black!65] (N) edge (M);
\end{tikzpicture}}
\caption{\label{fig:same_frame}
	Our model transfers argument roles of the verbal predicate ({\it acquired}) to the arguments of its nominalization ({\it acquisition}).
	\textit{acquired} and \textit{acquisition} share the lemma.
	Note that we do not rely on any argument alignment.
}
\vspace*{-.3cm}
\end{figure}


Most work on SRL relies on supervised learning~\citep{he2017deep,marcheggiani2017encoding},
and thus requires annotated resources such as PropBank~\citep{palmer2005proposition} and
FrameNet~\cite{baker1998berkeley} for English,
or SALSA~\cite{burchardt2006salsa} for German. However, the annotated data  is available only for a dozen of languages.
Moreover, many of these resources
cover only verbal predicates, whereas semantic relations are often triggered by other linguistic constructions, such as nominalizations and prepositions. For example, in the popular multilingual CoNLL-2009 dataset~\cite{hajivc2009conll}  nominal predicates are provided only for 3 languages (English, Czech, and Japanese). 
The scarcity of annotated data
motivated research into unsupervised SRL methods 
(e.g.,  
\citep{swier2004unsupervised,
	lang2011unsupervisedsplitmerge,
	titov2012bayesian,woodsend2015distributed}) 
but these approaches have also focused
only on verbal predicates, and, as we will show in our experiments, do not appear effective in the nominal SRL setting.
In contrast, 
nominal predicates, though often neglected in annotation efforts and model development,
have been shown crucial in many applications, such as extracting relations from scientific literature~\cite{bethard2008semantic} or events from social media~\cite{liu2012collective}.


In this work, we investigate a transfer scenario, where we assume the presence of role-annotated data for the `source' verbal domain but only unlabeled data for the `target' nominal domain. Our key assumption, driving the transfer between the two domains, is 
that selectional preferences of a role (i.e.,
preferences or constraints on the admissible arguments) do not strongly depend on whether the relation is triggered by a verb or a noun. 
Take the example in Figure~\ref{fig:same_frame}, semantically similar arguments can fill the A0 (`entity acquiring something') role for the verbal predicate {\it acquire} and its nominal form {\it acquisition}.


We approach the transfer problem from the generative perspective and use the variational autoencoding (VAE) framework~\cite{kingma2013auto} or, more specifically, its semi-supervised version~\cite{kingma2014semi}.
The semantic role labeler serves as an encoder (predicting roles given a sentence),
whereas selectional preferences are captured in the decoder component (generating arguments for the predicting roles). Nominal roles are not labeled in the training data, instead, the autoencoder's learning objective pushes the labeler to assign roles predictive of the arguments. Sharing the decoder parameters across the domains
encourages consistency between labels predicted for both domains.
Intuitively, the knowledge is transferred from the verbal domain 
to the nominal domain. 

Similarly, to work on unsupervised semantic role induction in the verbal domain
~\citep{swier2004unsupervised,lang2010unsupervised,titov2015unsupervised}, we focus solely on the argument labeling
subtask, and assume that candidate arguments are provided.
While we classify gold arguments in our experiments, candidate arguments 
can also be 
identified with simple rules~(see Section \ref{sec:auto_eval}). 

We experiment on the English CoNLL-2009 dataset and 
compare our approach to baselines. Among the baselines, we consider, (1) a direct transfer approach~\cite{zeman2008cross,sogaard2011data}, which  applies a verbal SRL model to the nominal data,
and (2) an unsupervised role induction method, which has been  shown to achieve state-of-the-art results on the verbal domain~\cite{titov2015unsupervised}.
We observe that our model outperforms the strongest baseline models by 6.58\% in accuracy and 4.27\% in F1, according to  supervised and unsupervised evaluation metrics, respectively. 
Our key contributions can be summarized as follows:
(\text{1}) we introduce a novel task of transferring SRL models from the verbal to the nominal domain, 
(2) we propose a simple latent variable model for the task, 
and (3) we show that the approach compares favourably to existing alternatives.

\section{Motivation}\label{sec:motivation}

\begin{table*}[!ht]
	\centering
	{\setlength{\tabcolsep}{.35em}
		\makebox[\linewidth]{\resizebox{\linewidth}{!}{%
				\begin{tabular}{ccP{0.325\linewidth}P{0.645\linewidth}}
					\toprule
					\toprule
					\multicolumn{4}{c}{\textbf{Most transferable predicate-role pairs}} \\
					\toprule
					\multirow{2}{*}{Predicate-Role} & \multirow{2}{*}{BC} & \multicolumn{2}{c}{Arguments} \\
					\cmidrule{3-4}
					&& Overlap & Complement (`non-overlap') \\
					\toprule
					\multirow{1}{*}{\makecell{earn \\  AM-TMP}} & \multirow{2}{*}{\makecell{0.845 \\ }} & 
					\Cpp{quarter}{0.44388}\Cpp{-num-}{0.14612}\Cpp{year}{0.14284}\Cpp{month}{0.02935}\Cpp{half}{0.02034}
					\Cpp{period}{0.01947}\Cpp{earlier}{0.01761}\Cpp{ago}{0.01017}\Cpp{annual}{0.00830}\Cpp{week}{0.00719} & 
					\MRpp{\textit{pretax}}{0.01840}\MBpp{season}{0.01685}\MBpp{time}{0.01685}\MBpp{month}{0.07022}\MRpp{\textit{interest}}{0.01227}\MBpp{current}{0.00843}\MBpp{night}{0.00843}\MBpp{now}{0.00843}\MBpp{yet}{0.00843}\MBpp{be}{0.00562}\MBpp{later}{0.00562}
					\MBpp{never}{0.00562}\MBpp{rise}{0.00562}\MBpp{year}{0.20787}\MBpp{still}{0.00562}\MBpp{summer}{0.00562}\MRpp{\textit{friday}}{0.00613}\MRpp{\textit{latest}}{0.00613}\MRpp{\textit{second}}{0.00613}\MRpp{\textit{tax}}{0.00613}\\
					\toprule
					\multirow{2}{*}{\makecell{increase \\  A2}} & \multirow{2}{*}{\makecell{0.821 \\ }} & 
					\Cpp{\%}{0.66517}\Cpp{\$}{0.09835}\Cpp{-num-}{0.02504}\Cpp{cent}{0.00946}\Cpp{more}{0.00946}\Cpp{million}{0.00669}
					\Cpp{share}{0.00669} &
					\MRpp{\textit{sharp}}{0.04138}\MBpp{mark}{0.02597}\MRpp{\textit{big}}{0.02069}\MRpp{\textit{small}}{0.02069}\MBpp{rate}{0.01948}\MRpp{\textit{huge}}{0.01379}\MRpp{\textit{low}}{0.01379}\MBpp{even}{0.01299}\MBpp{less}{0.01299}\MRpp{\textit{``}}{0.00690}\MRpp{\textit{dollar}}{0.00690}\MRpp{\textit{minor}}{0.00690}
					\MRpp{\textit{modest}}{0.00690}\MRpp{\textit{those}}{0.00690}\MRpp{\textit{total}}{0.00690}\MRpp{\textit{unit}}{0.00690}\MRpp{\textit{wage}}{0.00690}\MRpp{\textit{wide}}{0.00690}\MBpp{area}{0.00649}\MBpp{\$}{0.15584}	\\	
					\toprule
					\multirow{2}{*}{\makecell{sell \\  A1}} & \multirow{2}{*}{\makecell{0.732 \\ }} & 
					\Cpp{stock}{0.05702}\Cpp{share}{0.05550}\Cpp{asset}{0.05321}\Cpp{product}{0.04966}\Cpp{\$}{0.04608}
					\Cpp{bond}{0.03431}\Cpp{unit}{0.03403}\Cpp{car}{0.03140}\Cpp{stake}{0.02834}\Cpp{it}{0.01976} &
					\MRpp{\textit{retail}}{0.04092}\MBpp{them}{0.03084}\MRpp{\textit{truck}}{0.01790}\MRpp{\textit{auto}}{0.01535}\MBpp{yen}{0.01299}\MRpp{\textit{art}}{0.01023}\MBpp{amount}{0.00812}\MBpp{contract}{0.00812}\MBpp{million}{0.00812}\MRpp{\textit{gas}}{0.00767}\MRpp{\textit{glass}}{0.00767}
					\MRpp{\textit{part}}{0.01279}\MBpp{cd}{0.00568}\MBpp{debt}{0.00568}\MRpp{\textit{export}}{0.00512}\MRpp{\textit{lincoln}}{0.00512}\MBpp{record}{0.00487}\MRpp{\textit{fund}}{0.01790}\MRpp{\textit{car}}{0.06394}\MBpp{market}{0.00406} \\
					\toprule\toprule
					\multicolumn{4}{c}{\textbf{Least transferable predicate-role pairs}} \\
					\toprule
					\multirow{1}{*}{Predicate-Role} & \multirow{1}{*}{BC} &\multicolumn{2}{c}{Complement (`non-overlap')} \\
					\toprule
					\multirow{2}{*}{\makecell{interest \\  A2}} & \multirow{2}{*}{\makecell{0.028 \\ }} &  
					\MRpp{\textit{interest}}{0.33195}\MRpp{\textit{\%}}{0.09544}\MBpp{see}{0.12281}\MRpp{\textit{public}}{0.06224}\MRpp{\textit{minor}}{0.05809}\MRpp{\textit{investor}}{0.04564}\MBpp{buy}{0.07018}\MBpp{know}{0.07018}\MRpp{\textit{it}}{0.02905}\MRpp{\textit{his}}{0.02075}\MRpp{\textit{their}}{0.02075}\MRpp{\textit{creditor}}{0.01660}\MRpp{\textit{group}}{0.01660}\MRpp{\textit{holder}}{0.01660}\MRpp{\textit{u.s.}}{0.01660}\MBpp{push}{0.03509}\MRpp{\textit{buyer}}{0.01245}
					\MRpp{\textit{major}}{0.01245}\MRpp{\textit{our}}{0.01245}\MRpp{\textit{control}}{0.00830}\MRpp{\textit{half}}{0.00830}\MRpp{\textit{he}}{0.00830}\MRpp{\textit{jaguar}}{0.00830}\MRpp{\textit{nation}}{0.00830}\MRpp{\textit{p\&g}}{0.00830}\MRpp{\textit{third}}{0.00830}\MRpp{\textit{whose}}{0.00830}\MRpp{\textit{you}}{0.00830}\MBpp{make}{0.02339}\MBpp{thing}{0.02339}\\
					\toprule
					\multirow{2}{*}{\makecell{rate \\  A1}} & \multirow{2}{*}{\makecell{0.081 \\ }} &  
					\MRpp{\textit{interest}}{0.49654}\MBpp{film}{0.34431}\MBpp{stock}{0.16467}\MRpp{\textit{tax}}{0.10855}\MRpp{\textit{growth}}{0.07159}\MRpp{\textit{return}}{0.03464}\MBpp{show}{0.04940}\MRpp{\textit{loan}}{0.02771}\MRpp{\textit{default}}{0.02079}\MRpp{\textit{deposit}}{0.02079}\MRpp{\textit{discount}}{0.02079}\MRpp{\textit{bill}}{0.01848}\MRpp{\textit{profit}}{0.01848}\MBpp{him}{0.02844}\MRpp{\textit{ad}}{0.00924}\MRpp{\textit{death}}{0.00924}
					\MRpp{\textit{save}}{0.00924}\MRpp{\textit{\$}}{0.00693}\MBpp{debt}{0.02096}\MBpp{program}{0.01497}\MRpp{\textit{accept}}{0.00462}\MRpp{\textit{asset}}{0.00462}\MRpp{\textit{cd}}{0.00462}\MRpp{\textit{gain}}{0.00462}\MRpp{\textit{work}}{0.00462}\MBpp{share}{0.01347}\MBpp{them}{0.01198}\MBpp{he}{0.01048}\MRpp{\textit{bike}}{0.00231}\MRpp{\textit{cancer}}{0.00231}\\
					\toprule
					\multirow{2}{*}{\makecell{interest \\  A1}} & \multirow{2}{*}{\makecell{ 0.087 \\ }} &  
					\MRpp{\textit{interest}}{0.32639}\MBpp{me}{0.12000}\MRpp{\textit{short}}{0.07292}\MRpp{\textit{high}}{0.05208}\MRpp{\textit{\%}}{0.04861}\MRpp{\textit{buy}}{0.04167}\MRpp{\textit{low}}{0.03819}\MRpp{\textit{stock}}{0.03472}\MBpp{i}{0.07000}\MBpp{you}{0.06500}\MBpp{he}{0.05667}\MBpp{them}{0.05333}\MBpp{we}{0.05000}\MBpp{him}{0.04333}\MBpp{rate}{0.03500}\MBpp{investor}{0.03167}\MRpp{\textit{higher}}{0.02083}\MBpp{us}{0.02333}\MBpp{they}{0.01833}\MBpp{her}{0.01667}
					\MRpp{\textit{oil}}{0.01389}\MBpp{buyer}{0.01500}\MBpp{congressman}{0.01500}\MBpp{movement}{0.01333}\MBpp{other}{0.01333}\MBpp{team}{0.01333}\MBpp{voter}{0.01167}\MBpp{fan}{0.01000}\MBpp{itself}{0.01000}\MBpp{payment}{0.01000}\\
					\toprule
					\multirow{2}{*}{\makecell{result \\ A2}} & \multirow{2}{*}{\makecell{0.151 \\ }} &  
					\MRpp{\textit{result}}{0.73164}\MRpp{\textit{will}}{0.02542}\MRpp{\textit{may}}{0.01412}\MBpp{gain}{0.05133}\MRpp{\textit{have}}{0.04520}\MRpp{\textit{say}}{0.00847}\MBpp{cost}{0.03612}\MBpp{death}{0.03422}\MBpp{growth}{0.02852}\MRpp{\textit{remain}}{0.00565}\MRpp{\textit{would}}{0.00565}\MBpp{price}{0.02091}\MBpp{sale}{0.02091}\MBpp{rate}{0.01711}\MBpp{spend}{0.01711}\MBpp{\$}{0.01521}\MBpp{benefit}{0.01521}
					\MBpp{demand}{0.01521}\MBpp{lot}{0.01521}\MBpp{problem}{0.01521}\MBpp{it}{0.01331}\MRpp{\textit{ca}}{0.00282}\MRpp{\textit{can}}{0.00282}\MRpp{\textit{could}}{0.00282}\MRpp{\textit{direct}}{0.00282}\MRpp{\textit{do}}{0.00282}\MRpp{\textit{face}}{0.00282}\MRpp{\textit{fell}}{0.00282}\MRpp{\textit{fund}}{0.00282}\MRpp{\textit{make}}{0.00282}\\
					\toprule
					\multirow{2}{*}{\makecell{direct \\  A2}} & \multirow{2}{*}{\makecell{0.170 \\ }} &  
					\MRpp{\textit{direct}}{0.20859}\MBpp{buy}{0.13139}\MRpp{\textit{fund}}{0.03067}\MRpp{\textit{maker}}{0.03067}\MRpp{\textit{bureau}}{0.02454}\MBpp{be}{0.02555}\MBpp{make}{0.02555}\MBpp{state}{0.02555}\MBpp{veto}{0.02555}\MBpp{child}{0.02190}\MBpp{him}{0.02190}\MBpp{review}{0.02190}\MRpp{\textit{\&}}{0.01840}\MRpp{\textit{brother}}{0.01840}\MRpp{\textit{co.}}{0.01840}\MRpp{\textit{motor}}{0.01840}\MRpp{\textit{unit}}{0.01840}\MBpp{``}{0.01825}
					\MBpp{me}{0.01825}\MBpp{attorney}{0.01460}\MBpp{tv}{0.01460}\MRpp{\textit{america}}{0.01227}\MRpp{\textit{australia}}{0.01227}\MRpp{\textit{budget}}{0.01227}\MRpp{\textit{firm}}{0.01227}\MRpp{\textit{for}}{0.01227}\MRpp{\textit{hutton}}{0.01227}\MRpp{\textit{research}}{0.01227}\MRpp{\textit{right}}{0.01227}\MRpp{\textit{trust}}{0.01227}\\
					\toprule
					\multirow{2}{*}{\makecell{order \\  A1}} & \multirow{2}{*}{\makecell{0.176 \\ }} &  
					\MRpp{\textit{order}}{0.22378}\MRpp{\textit{buy}}{0.11189}\MRpp{\textit{sell}}{0.07692}\MBpp{he}{0.05680}\MRpp{\textit{obtain}}{0.02797}\MRpp{\textit{ban}}{0.02098}\MBpp{him}{0.03448}\MBpp{worth}{0.02840}\MBpp{more}{0.02637}\MBpp{court}{0.02231}\MRpp{\textit{be}}{0.01399}\MRpp{\textit{model}}{0.01399}\MRpp{\textit{prevent}}{0.01399}\MRpp{\textit{good}}{0.13287}\MRpp{\textit{stop}}{0.01399}\MBpp{her}{0.01826}\MBpp{them}{0.01826}\MBpp{bank}{0.01623}
					\MBpp{report}{0.01623}\MRpp{\textit{allow}}{0.00699}\MRpp{\textit{back}}{0.00699}\MRpp{\textit{block}}{0.00699}\MRpp{\textit{bond}}{0.00699}\MRpp{\textit{car}}{0.00699}\MRpp{\textit{center}}{0.00699}\MRpp{\textit{extend}}{0.00699}\MRpp{\textit{find}}{0.00699}\MRpp{\textit{gain}}{0.00699}\MRpp{\textit{gap}}{0.00699}\MRpp{\textit{get}}{0.00699}\\
					\toprule
					\multirow{2}{*}{\makecell{make \\  A1}} & \multirow{2}{*}{\makecell{0.233 \\ }} &  
					\MRpp{\textit{auto}}{0.29520}\MBpp{it}{0.07621}\MBpp{sure}{0.03976}\MBpp{them}{0.03844}\MBpp{offer}{0.03115}\MBpp{loan}{0.01988}\MBpp{move}{0.01789}\MBpp{statement}{0.01723}\MBpp{payment}{0.01590}\MRpp{\textit{printer}}{0.01476}\MBpp{\$}{0.01524}\MBpp{effort}{0.01392}\MBpp{be}{0.01193}\MBpp{him}{0.01193}\MBpp{case}{0.01127}\MBpp{lot}{0.01127}\MBpp{me}{0.01127}\MBpp{invest}{0.01060}
					\MBpp{point}{0.00928}\MBpp{thing}{0.00928}\MBpp{you}{0.00928}\MBpp{deposit}{0.00795}\MBpp{progress}{0.00795}\MBpp{us}{0.00795}\MBpp{\%}{0.00729}\MBpp{cut}{0.00729}\MBpp{sale}{0.00729}\MBpp{trip}{0.00729}\MBpp{use}{0.00729}\MBpp{comment}{0.00663}\\
					\bottomrule
					\bottomrule
	\end{tabular}}}}
	\vspace*{-.1em}
	\caption{\label{tab:most_and_least} 
		Predicate-role pairs with the most and least similar selectional preferences, as
		measured with the Bhattacharyya coefficient (BC),  the larger the more similar.
		The intensity of green is used to indicate the degree of positive contribution of an argument to BC.
		The intensity of blue and red indicates the negative contribution:
		the blue arguments appear predominantly with  verbal predicates,
		where the red (= italic) words are mostly arguments of nominalizations.
	}
\vspace*{-.5cm}
\end{table*}

As discussed above, our approach
hinges on the assumption that selectional preferences
of a role (e.g., A0) are similar for verbal predicates (e.g., {\it acquire}) and their nominalizations (e.g., {\it acquisition}). In this section, we verify 
this assumption through analyzing  the verbal PropBank~\citep{palmer2005proposition} and nominal NomBank~\citep{meyers2004nombank} resources, or, more formally, their CoNLL-2009 version \citep{hajivc2009conll}. 

As in the rest of the paper, we rely
on lemmas and a nominalization list\footnote{\url{http://amr.isi.edu/download/lists/verbalization-list-v1.06.txt}} 
to relate verbal predicates (e.g., {\it earn}) to their
nominal forms (e.g., {\it earning} and {\it earner})\footnote{Note that their roles
	may not be fully aligned, even though NomBank annotators
	were encouraged to reuse PropBank frames~\cite{meyers2004nombank}. Also, we ignore predicate senses (i.e., `sense set' identifiers) in this work.}. 
In what follows, we use lemmas to refer to predicates. 
To ensure the reliability of our metrics, we consider predicate-role pairs which appear at least 100 times, 
resulting in 65 predicate-role pairs, and take only frequent arguments (a cut-off of 20). 

First, we would like to identify the predicate-role pairs which have the most and the least similar selectional preferences across the domains. To do this, for every predicate-role pair, we collect its arguments in the verbal domain and arguments in the nominal domain, lemmatize them and measure the distance between the two argument samples. We rely on the Bhattacharyya coefficient (BC), a standard way of measuring the amount of overlap between statistical samples. As we can see from Table~\ref{tab:most_and_least}, BC for ({\it earn}, AM-TMP) and ({\it increase}, A2) are the highest ($0.845$ and $0.821$), indicating that the selectional preferences
for the nominal and verbal domains are very similar. In contrast, the similarity is very 
low ($0.028$ and $0.081$) for ({\it interest}, A2) and ({\it rate}, A1). Arguments
for these role-predicate pairs almost do not overlap between their verbal and nominal forms.

Second, in order to understand the reason for this discrepancy, we measure the contribution of individual arguments. 
We define the contribution as the change
to BC, caused by removing all occurrences of the argument from the samples.
For example, removing {\it quarter} for the predicate-role pair ({\it earn}, AM-TMP)
causes the largest drop in BC, while taking out {\it pretax} results in the largest jump.
We can see that some of the largest discrepancies are due to peculiarities of the NomBank annotation guidelines. The most obvious one is  the presence of {\it self-loops}, i.e., nominal predicates in NomBank can take themselves as arguments. 
For example, A2 for the predicate {\it direct} is defined as `movement direction',
and the nominalized predicate {\it direction} is always labeled as A2. 
Other problematic cases are due to  multi-word expressions, such as {\it auto makers} or {\it pretax earnings} that do not have frequent verbal counterparts.

Still, even if we ignore annotation divergences, we see that our assumption is only an approximation. Nevertheless, it is accurate  for a subset of predicate-role pairs. We hypothesize that, due to parameter sharing in our model, easier `transferable' cases will help us to learn to label in harder `non-transferable' ones.

\section{Transfer Learning}

In our setting, we are provided with a labeled dataset $\mathcal{X}_v$ for the source verbal domain but the role labels are not annotated in the target nominal data $\mathcal{X}_n$.   
Our goal is to produce an SRL model for the target.
Given a predicate $p$ in a sentence $\mb{w}$, a semantic role labeler needs to assign roles $\mb{y} = y_1, \ldots, y_m$  to arguments $\mb{a} = a_1, \ldots, a_m$. We consider dependency-based SRL~(see Figure~\ref{fig:same_frame}), i.e., we label syntactic heads of argument phrases (e.g., `positions' instead of the entire span `equity positions').

In order to exploit unlabeled data we will use the generative framework, or, more specifically,
semi-supervised VAEs~\citep{kingma2014semi}. 
We will start by describing 
the generative model, and then show how it can be integrated in the semi-supervised VAE objective and used to produce a discriminative semantic role labeler.

\subsection{Generative model}\label{sec:vae_generative_model}

The generative model defines the process of producing arguments $\mb{a}$ relying on roles $\mb{y}$ and a continuous latent variable $\mb{z}$. The variable $\mb{z}$ can be regarded as a latent code encoding properties of the proposition (i.e., the triple $p$, $\mb{y}$ and $\mb{a}$).
We start by drawing the roles and the code from priors:
\begin{align}
& p(\mb{y}) = \prod_i p(y_i); \quad  p(\mb{z}) = \mathcal{N}(\mb{z} | \mb{0}, \mb{I}),
\nonumber
\end{align}
where, for simplicity, we assume that $p(y_i)$ is the uniform distribution. We then generate arguments (formally, argument lemmas)
using an autoregressive selectional preference model:
\begin{align}
p_{\theta}(\mb{a} | \mb{y}, \mb{z})
= \prod_i{ p_{\theta}(a_i| \mb{y}, \mb{a}_{<i}, \mb{z}, p) },
\end{align}
where $\mb{a}_{<i} = (a_1, \ldots, a_{i-1})$.
This contrasts with most previous work
in selectional preference modeling, which assumes
that arguments are conditionally-independent  (e.g., \cite{ritter2010latent}). As we will see in our ablations, both using the latent code and conditioning
on other arguments is beneficial.
On nominal data $\mb{y}$ is latent 
and the marginal likelihood is given by
\begin{flalign}
\label{eq:ll}
T_\theta(\mb{a})\! = \!
\log \! \int \sum_{\mb{y}} 
p(\mb{y})  p(\mb{z}) 
p_{\theta}(\mb{a} | \mb{y}, \mb{z})
\,d\mb{z}\,.
\end{flalign}
Recall that
we reuse the generative model across nominal and verbal domains. As the preferred argument order
is different for verbal and nominal predicates, our model needs to be permutation-invariant or, equivalently, model bags of labeled arguments rather than their sequences~\cite{vinyals2015order}.
We achieve this by replacing the maximum likelihood objective $T_\theta(\mb{a})$ with pseudolikelihood~\cite{besag1977efficiency}. In other words,
we replace $p_{\theta}(\mb{a} | \mb{y}, \mb{z})$ in Equation~\ref{eq:ll} with the following term:
\begin{align}
f_\theta(\mb{a}, \mb{y}, \mb{z}) = \prod_i{ p_{\theta}(a_i| \mb{y}, \mb{a}_{-i}, \mb{z}, p)},  
\end{align}
where $\mb{a}_{-i}$ refers to all arguments in $\bf{a}$ but the $i$-th, resulting
in the objective
\begin{align}
\hat{T}_\theta(\mb{a})\! = \!
\log \! \int \sum_{\mb{y}} 
p(\mb{y})  p(\mb{z}) 
f_\theta(\mb{a}, \mb{y}, \mb{z})
\,d\mb{z}\,.  
\end{align}
Intuitively,
when predicting an argument, the model can access all other arguments.
Furthermore, to fully get rid of the order information, we ensure that $p_{\theta}(a_i| \mb{y}, \mb{a}_{-i}, \mb{z}, p)$ is invariant to joint reordering of the arguments $\mb{a}_{-i}$ and their roles $\mb{y}_{-i}$. The neural network achieving this property
is described in Section~\ref{sec:vae_c}.

\subsection{Argument reconstruction}\label{sec:gen_vae}

Optimizing the pseudolikelihood involves intractable marginalization over the latent variables $\mb{z}$ and $\mb{y}$.
With VAEs, we instead maximize its lower bound:
\begin{flalign}\label{eq:gen_unlabeled}
\hat{T}_{\theta}(\mb{a}) &\ge 
\mathbb{E}_{q_{\phi}(\mb{y}, \mb{z} | \mb{w})}\left[
\log f_\theta(\mb{a}, \mb{y}, \mb{z}) \right]  \\
&-\text{KL}(q_{\phi}(\mb{y},\mb{z} | \mb{w}) || p(\mb{y})p(\mb{z}))  = \mathcal{U}_{\theta,\phi}(\mb{w}) \,,\nonumber
\end{flalign}
where $q_{\phi}(\mb{y}, \mb{z} | \mb{w})$ is
an estimate of the intractable
true posterior,
provided by an {\it inference network}, a neural network parameterized by $\phi$.
The first term in Equation~\ref{eq:gen_unlabeled} 
is the autoencoder's reconstruction loss:  the inference network serves as an encoder and the selectional preference model as the decoder. The other term can be regarded as a regularizer on the latent space.   
We also factorize
the inference network into two components: a semantic role labeler $q_{\phi}(\mb{y} | \mb{w})$
and a latent code generator $q_{\phi}(\mb{z} | \mb{w})$.

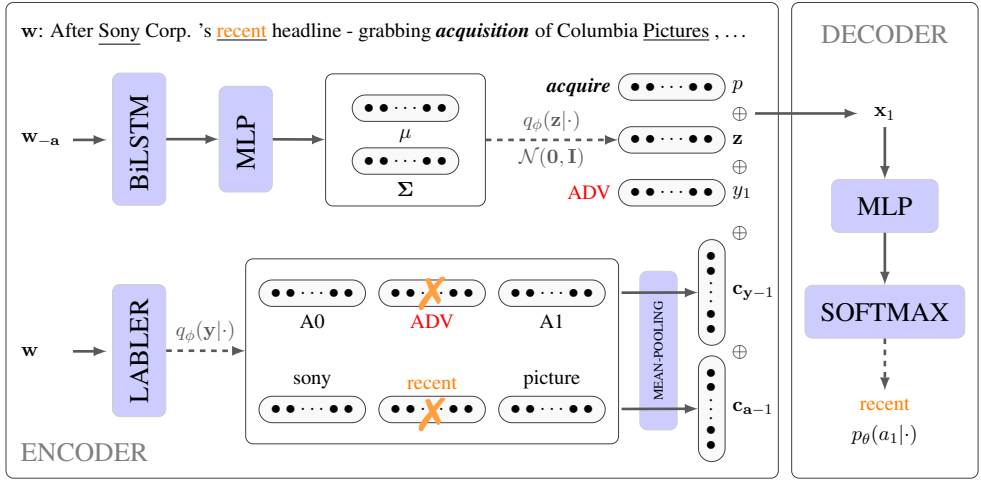
\begin{figure*}
\centering
\resizebox{.8\linewidth}{!}{%
\begin{tikzpicture}[node distance=0mm]
\node[annotation,draw=none] (Sentence) at (0.15, 15.25) {$\mb{w}$: After \underline{Sony} Corp. ’s \underline{\argument{recent}} headline - grabbing \textbf{\textit{acquisition}} of Columbia \underline{Pictures} , …};
\node[annotation,draw=none,minimum width=11mm,align=left] (WA) at (0.15, 13.25) { $\mb{w}_{-\mb{a}}$ };
\node[vmodule,minimum width=25mm] (BiLSTM) at (2, 11.5) { BiLSTM };
\node[vmodule,minimum width=20mm] (VMLP) at (4, 11.75) { MLP };
\node[rounded corners,anchor=north west,minimum width=10mm,minimum height=25mm] (REC11) at (1.9, 14) {};

\node[hvector,label=below:{$\mb{\mu}$}] (V11) at (6.5, 13.6) {$\bullet\bullet\cdots\bullet\bullet$};
\node[hvector,label=below:{$\mb{\Sigma}$}] (V12) at (6.5, 12.6) {$\bullet\bullet\cdots\bullet\bullet$};

\node[rounded corners,draw=black!75,anchor=north west,minimum width=30mm,minimum height=25mm] (REC1) at (6, 14) {};

\node[hvector,label=left:{\textbf{\textit{acquire}}},label=right:{$p$}] (V21) at (11.5, 14) {$\bullet\bullet\cdots\bullet\bullet$};
\node[hvector,label=right:{$\bz$}] (V22) at (11.5, 13) {$\bullet\bullet\cdots\bullet\bullet$};
\node[hvector,label=left:{\role{ADV}},label=right:{$y_1$}] (V23) at (11.5, 12) {$\bullet\bullet\cdots\bullet\bullet$};
\node[annotation,align=left,draw=none,minimum width=11mm,text width=8mm] (W) at (0.15, 9.25) {$\mb{w}$};
\node[vmodule,minimum width=25mm] (LABLER1) at (2, 7.5){LABLER};
\node[rounded corners,anchor=north west,minimum width=10mm,minimum height=25mm] (REC12) at (1.9, 10) {};

\node[hvector,label=below:{A0}] (V31) at (4.75, 10.1) {$\bullet\bullet\cdots\bullet\bullet$};
\node[hvector,label=below:{\role{ADV}}] (V32) at (7, 10.1) {$\bullet\bullet\cdots\bullet\bullet$};
\node[hvector,label=below:{A1}] (V33) at (9.25, 10.1) {$\bullet\bullet\cdots\bullet\bullet$};

\node[rounded corners, align=center, anchor=north west, font=\Huge, 
minimum width=20mm] (ENCODER) at (7, 10.3){\textcolor{orange!75}{\textbf{\xmark}}};

\node[hvector,label=above:{sony}] (V43) at (4.75, 7.9) {$\bullet\bullet\cdots\bullet\bullet$};
\node[hvector,label=above:{\argument{recent}}] (V43) at (7, 7.9) {$\bullet\bullet\cdots\bullet\bullet$};
\node[hvector,label=above:{picture}] (V43) at (9.25, 7.9) {$\bullet\bullet\cdots\bullet\bullet$};

\node[rounded corners, align=center, anchor=north west, font=\Huge, 
minimum width=20mm] (ENCODER) at (7, 8.0){\textcolor{orange!75}{\textbf{\xmark}}};

\node[rounded corners,draw=black!75,anchor=north west,minimum width=70mm,minimum height=35mm] (REC2) at (4.5, 10.5) {};
\node[rounded corners,draw=none,anchor=north west,minimum width=70.5mm,minimum height=13mm] (REC21) at (4.5, 10.5) {};
\node[rounded corners,draw=none,anchor=north west,minimum width=70.5mm,minimum height=13mm] (REC22) at (4.5, 8.3) {};

\node[vmodule,minimum width=30mm,minimum height=7mm] (MEAN) at (11.9, 7.25) {\scriptsize{MEAN-POOLING}};

\node[vvector,label=below:{$\mb{c}_{\mb{y}-1}$}] (V51) at (13, 8.85) {$\bullet\bullet\cdots\bullet\bullet$};
\node[vvector,label=below:{$\mb{c}_{\mb{a}-1}$}] (V52) at (13, 6.65) {$\bullet\bullet\cdots\bullet\bullet$};
\node[annotation,draw=none,minimum height=5mm] (X) at (16, 13.5) {$\mb{x}_1$};
\node[hmodule,minimum width=20mm] (HMLP) at (15.5, 12){MLP};
\node[hmodule,minimum width=30mm] (SM) at (15, 10){SOFTMAX};
\node[annotation,draw=none,minimum height=5mm,minimum width=15mm,label=below:{$p_{\theta}(a_1 | \cdot)$}] (ARG) at (15.75, 8) {\argument{recent}};
\node[annotation,minimum width=5mm,minimum height=7mm,draw=none] (OP1) at (13.5, 13.6) {$\oplus$};
\node[annotation,minimum width=5mm,minimum height=5mm,draw=none] (OP2) at (13.5, 12.5) {$\oplus$};
\node[annotation,minimum width=5mm,minimum height=5mm,draw=none] (OP3) at (13.5, 11.25) {$\oplus$};
\node[annotation,minimum width=5mm,minimum height=5mm,draw=none] (OP4) at (13.5, 9) {$\oplus$};

\node[rounded corners,draw=black!75,anchor=north west,minimum width=145mm,minimum height=90mm] (REC3) at (0, 15.35) {};
\node[rounded corners,draw=black!75,anchor=north west,minimum width=35mm,minimum height=90mm] (REC4) at (14.75, 15.35) {};

\node[rounded corners, align=center, anchor=north west, font=\Large, 
minimum height=10mm, minimum width=30mm] (DECODER) at (15, 15.25){\textcolor{gray}{DECODER}};
\node[rounded corners, align=center, anchor=north west, font=\Large, 
minimum height=10mm] (ENCODER) at (0.15, 7.35){\textcolor{gray}{ENCODER}};
\draw [line] (WA) -- (REC11);
\draw [line] (BiLSTM) -- (VMLP);
\draw [line] (VMLP) -- (REC1);
\draw [line] (REC1) -- (V22) [dashed] node[pos=.5,below] {$\mb{\mathcal{N}}(\mb{0}, \mb{I})$} node[pos=.5,above] {$q_{\phi}(\mb{z}|\cdot)$};

\draw [line] (W) edge (REC12);
\draw [line] (LABLER1) -- (REC2) [dashed] node[pos=.5,above] {$q_{\phi}(\mb{y}|\cdot)$};
\draw [line] (REC21) -- (V51);
\draw [line] (REC22) -- (V52);

\draw [line] (X) -- (HMLP);
\draw [line] (HMLP) -- (SM);
\draw [line] (SM) -- (ARG) [dashed];

\draw [line] (OP1) -- (X);

\end{tikzpicture}}
	\caption{\label{fig:model_overview}
	An example of predicting the argument \textit{\textbf{recent}} with the VAE model. 
	The underlined words in $\mb{w}$ are argument fillers.
	We replace them with a special symbol '\_' and obtain $\mb{w}_{-\mb{a}}$.
	$\mb{x}_1$ is the input to the decoder model.
	It is represented as the concatenation of the predicate (\textit{acquire}) embedding, the latent code $\mb{z}$, the embedding of the \textit{\textbf{recent}}'s role (ADV), 
	and contexts from the other roles and arguments.
	Note that role labels are sampled from $q_{\phi}(\y|\cdot)$;
	we are using lemmas for the nominal predicate \textit{acquisition} and  the arguments.}
\vspace*{-.5cm}
\end{figure*}


On the verbal data  $\mb{y}$ is observed, and the pseudo-likelhood lower bound  becomes:
\begin{flalign}\label{eq:gen_labeled}
\mathcal{L}_{\theta,\phi}(\mb{w}, \mb{y}) & =
\mathbb{E}_{q_{\phi}(\mb{z} | \mb{w})}\left[
\log f_{\theta}(\mb{a}, \mb{y}, \mb{z}) \right]  \\
&-\text{KL}(q_{\phi}(\mb{z} | \mb{w}) || p(\mb{z})) \,.\nonumber
\end{flalign}
We will describe the architecture of both inference networks in Section~\ref{sec:design}.

\subsection{Joint objective}\label{sec:vae_transfer}

While using the objectives $\mathcal{U}_{\theta,\phi}(\mb{w})$ 
and $\mathcal{L}_{\theta,\phi}(\mb{w},\mb{y})$ can already faciliate the transfer, 
as standard with semi-supervised VAEs, we use an extra discriminative term
which encourages the semantic role labeler to predict
correct labels on the labeled verbal data:
\begin{flalign}
\label{eq:sup-loss}
\mathcal{D}_{\phi}(\mb{w}_{v}, \mb{y}_{v}) = \log q_{\phi}(\mb{y}_{v} | \mb{w}_{v})\,.
\end{flalign}

In practice, we  share 
the semantic role labeler $q_{\phi}(\mb{y} | \mb{w})$ across nominal and verbal domains.
Note that though the parameters $\phi$ are shared, the
labeler relies on the predicate and the sentence and can easily detect which domain it is applied to. In other words,  sharing the labeler's parameters provides only a soft inductive bias rather than a hard constraint.
The parameter sharing is the reason for why $q_\phi(\mb{y}_{v} | \mb{w}_{v})$ can benefit from accessing the entire sentence $\bf{w}$, rather than using only the arguments $\bf{a}$, as would have followed from the posterior estimation perspective.

To summarize, the overall loss function is minimized with respect to $\theta$ and $\phi$:
\begin{flalign}
\label{eq:final-obj}
\mathcal{J_{\theta,\phi}} &= 
- \sum_{\mb{w}_n \sim \mathcal{X}_n} \mathcal{U}_{\theta,\phi}(\mb{w}_n) \\
&-\sum_{(\mb{w}_v, \mb{y}_v) \sim \mathcal{X}_v}  \mathcal{L}_{\theta,\phi}(\mb{w}_v, \mb{y}_v) + \alpha \cdot \mathcal{D}_{\phi}(\mb{w}_v, \mb{y}_v)\,,   \nonumber 
\end{flalign}
where $\alpha$ is a hyper-parameter balancing the relative importance of the discriminative term.
Note that the estimation of the lower bounds requires sampling from  $q_{\phi}(\mb{y} | \mb{w})$ and $q_{\phi}(\mb{z} | \mb{w})$,
which is not differentiable. 
This, together with the presence of discrete latent variables, makes the optimization challenging.
As a remedy,  
for $\mb{z}$ we employ the reparametrization trick~\citep{kingma2014semi};
for $\mb{y}$ we resort to the Gumbel-Softmax relaxation~\citep{jang2016categorical,maddison2016concrete}.
Details can be found in Appendix~\ref{app:supp_learning}.

\section{Component Architecture}
\label{sec:design}

In this section, we formally describe every component of our VAE.
A schematic overview of the model is illustrated in Figure~\ref{fig:model_overview}.

\subsection{Decoder $f_\theta(\mb{a}, \mb{y}, \mb{z})$}
\label{sec:vae_c}

As shown on the right in Figure~\ref{fig:model_overview}, the neural network computing
$p_{\theta}(a_i| \mb{y}, \mb{a}_{-i}, \mb{z}, p)$
takes as input
the concatenation of the predicate embedding $p$, 
the latent code $\mb{z}$, the role embedding $y_i$ as well as
the representations of the other arguments $\mb{c}_{\mb{a}_{-i}}$ and the other roles $\mb{c}_{\mb{y}_{-i}}$:
\begin{flalign}
\mb{c}_{\mb{a}_{-i}} &= \frac{1}{m - 1} \sum_{j\neq i} 
\mb{v}_{a_j}\,, \nonumber\\
\mb{c}_{\mb{y}_{-i}} &= \frac{1}{m - 1} \sum_{j\neq i} 
\mb{v}_{y_{j}}\,, \nonumber
\end{flalign}
where $\mb{v}_{a_j}$ and $\mb{v}_{y_{j}}$ are argument embedding and role embedding, respectively.
The resulting vector is passed through a two-layer feedforward network, with the {\it tanh}
non-linearities, followed with the softmax over argument lemmas.
The arguments $\mb{a}_{-i}$ and their roles $\mb{y}_{-i}$
are encoded via the mean operation over their embeddings, ensuring that the resulting model
is invariant to their order.

\subsection{Encoder models}\label{sec:vae_encoder}

\subsubsection{Semantic role labeler $q_{\phi}(\mb{y} | \mb{w})$}\label{sec:vae_y}

We adapt the SRL model of~\citet{he2017deep} to dependency-based SRL.
Specifically, 
a highway BiLSTM model first encodes arguments into contextualized representations.
Then roles are predicted independently using the softmax function.
Note that the SRL model is the only component retained after training 
with the objective ~(defined in Equation \ref{eq:final-obj}) and used at test time. Other model components
can be discarded.

\subsubsection{Latent code predictor $q_{\phi}(\mb{z} | \mb{w})$}\label{sec:vae_z}

Formally, as the inference network estimates
the posterior $p(\mb{z} | \mb{a})$, it needs to rely on all arguments $\mb{a}$
and cannot benefit from any other information in $\mb{w}$. In practice, using
such an inference network would result in {\it posterior collapse}: 
the model would ignore discrete variables $\mb{y}$ and rely on $\mb{z}$
to pass information to the decoder. In order to prevent
this,  the network
relies on $\mb{w}_{-\mb{a}}$, i.e., the sentence with all the predicate's arguments masked.
In other words, it can derive relevant features only from the context.
In our implementation, we input $\mb{w}_{-\mb{a}}$ to a BiLSTM model,
then a mean-pooling layer is applied to the BiLSTM hidden states.
Finally, the mean vector $\mb{\mu}_{\phi}(\mb{w}_{-\mb{a}})$ and the log-covariances $\log \mb{\sigma}_{\phi}(\mb{w}_{-\mb{a}})$ are output by an affine layer. 

\section{Experiments}

The main results are presented in Section~\ref{sec:supervised_eval}.
As unsupervised baselines cannot be evaluated using standard metrics,
we compare with them using clustering metrics in Section~\ref{sec:unsupervised_eval}.

\subsection{Datasets and model configurations}\label{sec:exp_data}

\noindent{\textbf{Dataset}}: 
We use the English CoNLL-2009 (CoNLL) dataset~\citep{hajivc2009conll} with standard data splits.  
We separate it into the verbal and nominal domains by predicate types: verb vs noun.
For a prepositional argument such as \textit{at} in the prepositional phrase `\textit{at the library}', 
we instead use as the argument the headword (\textit{library} in this case) of the noun phrase following the preposition.
Moreover, we keep 15 frequent roles shared by the two domains (a cut-off of 3).
Data statistics and processing details can be found 
in Appendix~\ref{app:data_conll}. 

\noindent{\textbf{Data augmentation}}:
Transfer learning models benefit from labeled data in the source domain. 
To obtain extra labeled verbal data,
we pre-train a verbal semantic role labeler 
and apply it to the verbal predicates in the New York Times Corpus (NYT)~\citep{sandhaus2008new}
(details in Appendix~\ref{app:data_augmentation}).
For each predicate in the unlabeled nominal training data, we select 1000 random instances with that predicate from the automatically-labeled NYT data and add them to our verbal training data.

\noindent{\textbf{Model configurations}}: 
Our model is implemented using AllenNLP~\citep{gardner2018allennlp}
and relies on ELMo~\citep{peters2018deep}.
The dimension of the latent code $\mb{z}$ is set to 100.
The embeddings of predicates and roles are randomly initialized.
We pre-train lemma embeddings for arguments on the Wikipedia data~\citep{muller2015robust},
which is lemmatized by the lemmatizer~\textsc{Lemming}~\citep{muller2015joint}.
All the embeddings are of dimension 100 and are tunable during training.
We optimize the model using the Adadelta~\citep{zeiler2012adadelta},
with the learning rate set to 1.
Appendix~\ref{app:model_params} provides extra details.

\subsection{Automatic argument identification}\label{sec:auto_eval}

We start by demonstrating that unlabeled arguments can be extracted with relatively simple rules. 
Inspired by heuristics from \citet{lang2011unsupervisedsplitmerge} developed for verbal predicates,
we define 5 simple  rules; these rules use syntactic dependency structures and yield 77\% F1 in identification. 
Appendix~\ref{app:heuristics} elaborates on the rules.
The distribution of labels of the extracted arguments is however different from the one in gold standard.
E.g., A0 is almost twice more frequent among gold arguments than extracted ones.
Thus, to make our findings
 independent of the (imperfect and simple) rules, in main experiments we focus on classifying arguments taken from ground truth. 

\subsection{Supervised evaluation}\label{sec:supervised_eval}

\subsubsection{Baseline models}\label{sec:supervised_baselines}

We consider the following three baseline models:
\begin{itemize}[wide=0\parindent,noitemsep]
	\item[1.]\text{\it Most-frequent} assigns an argument to its most frequent role given the predicate in the verbal training data.
	When the  predicate and the argument do not co-occur, 
	it assigns the argument to the most frequent role of the predicate.
	\item[2.]\text{\it Factorization} 
	estimates the compatibility of a role $y_i$ with predicate $p$ and argument $a_i$ using 
	the bilinear score
	$s(a_i, y_i, p) = \mb{v}_{a_i}^{T}\mb{W}_{y_i}\mb{v}_p + \mb{v}_{a_i}^{T}\mb{w}_{y_i}$,
	where $\mb{v}_{a_i}$ and $\mb{v}_p$ are argument and predicate embeddings, respectively; 
	$\mb{W}_{y_i}$ and $\mb{w}_{y_i}$ are a matrix embedding and a vector embedding of the role, respectively. 
	The scoring function is similar
	to the one used in ~\citet{titov2015unsupervised} but we estimate it on the labeled verbal data.
	See Appendix~\ref{app:sup_bl} for details. 	
	\item[3.]\text{\it Direct-transfer} is our semantic role labeler estimated only on the verbal data
	(i.e., using the loss $\mathcal{D}_\phi$ in Equation ~\ref{eq:sup-loss}) and applied to the nominal data.
\end{itemize}

\begin{table}[t!]\small
	\renewcommand{\arraystretch}{1.00}
	\centering
	\vskip .0in
	{\setlength{\tabcolsep}{.45em}
		\makebox[\linewidth]{\resizebox{\linewidth}{!}{%
				\begin{tabular}{rrrrrr}
					\toprule
					Model & A0 & A1 & A2 & AM-* & \textit{All} \\
					\toprule
					Most-frequent & 62.45 & 67.13 & 17.69 & 42.11 & 56.51\\
					Factorization & 53.76 & 56.45 & 3.65 & 27.03 & 44.48 \\
					Direct-transfer & 67.52 & 64.72 & 24.13 & 40.78 & 55.85 \\
					Ours & 71.43 & 70.09 & 19.78 & 62.04 & \textbf{63.09} \\
					\toprule
					-Self-loop & 66.44 & 73.80 & 27.26 & 63.07 & 64.51 \\
					\bottomrule
	\end{tabular}}}}
	\vspace*{-.5em}
	\caption{\label{tab:supervised_results} F1 scores on the CoNLL nominal test data.}
	\vspace*{-.1cm}
\end{table}	

\begin{figure}
	\begin{center}
		\includegraphics[width=1.0\linewidth]{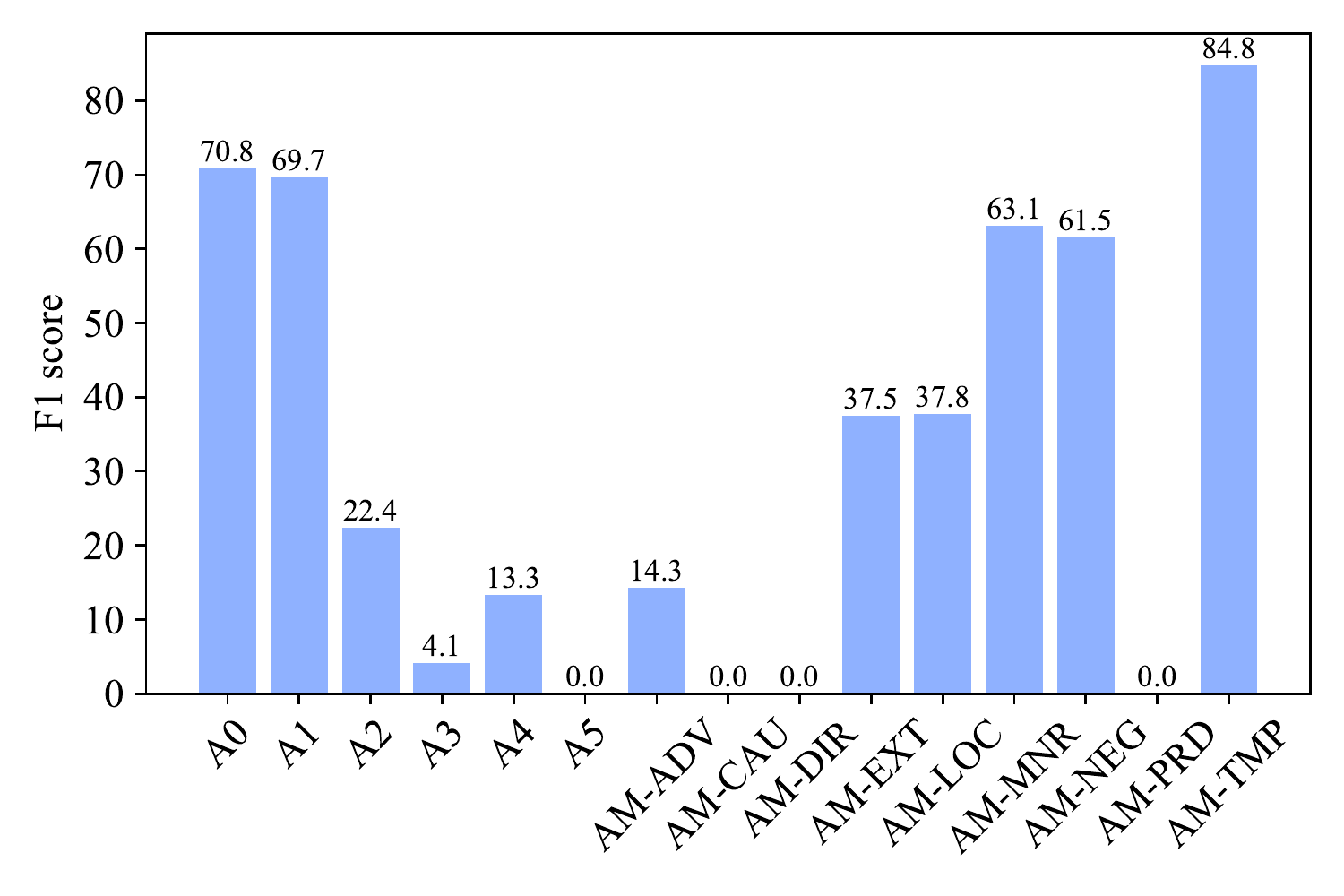}
	\end{center}
	\vspace*{-1.0em}
	\caption{\label{fig:f1_roles}F1 score of our model, per role, on the nominal devel data.}
	\vspace*{-.5cm}
\end{figure}

\subsubsection{Experimental results}\label{sec:supervised_results}

Our model achieves the best overall scores (Table~\ref{tab:supervised_results}).
{\it Most-frequent} and {\it Factorization} directly estimate selectional preferences
on the verbal domain and apply them to the nominal one, 
while in our approach we share a selectional preference model to bootstrap our labeler. The lower scores of these baselines suggest that it is indeed crucial to distill this weak signal into the SRL model rather than relying on it at test time. The improvement over {\it Direct-transfer} shows that we genuinely benefit from using unlabeled nominal data.

The performance varies greatly across the roles, with the best
results for proto-agent (A0) and proto-patient (A1) roles
and much weaker for A2-A5 and adjunt / modifier roles (AM-*).
Roles A2-A5 are hard even for supervised systems, as they are predicate-specific
(i.e., A4 for one predicate may have nothing to do with A4 for another).
The low scores for AM-* are more surprising,  
as their selectional preferences should not depend much on the predicate. 
Looking into individual roles in Figure~\ref{fig:f1_roles},
we see that our model is accurate for AM-TMP, but not
for AM-ADV and AM-MNR. 
While for AM-TMP lemmas of arguments are very similar
across verbal and nominal domains (see also top line in Table~\ref{tab:most_and_least}),
this is not the case for AM-ADV and AM-MNR.
E.g., while frequent AM-ADV arguments in the verbal domain are {\it so} and 
{\it while},  these lemmas do not  appear as AM-ADV arguments in the nominal data.  


The nominal data contains self-loops, i.e., cases where a predicate is an argument of itself  (see Section~\ref{sec:motivation}). This
annotation decision is controversial and most other formalism~\cite{banarescu2013abstract} 
does not use them. 
Their presence negatively affects our approach as the verbal data has no self-loops.
As expected, our model  scores better when these arguments are removed at test time (\textit{-Self-loop} in Table~\ref{tab:supervised_results}). 

\begin{table}[t!]
	\centering
	\vskip .0in
	\renewcommand{\arraystretch}{1.0}
	{\setlength{\tabcolsep}{.30em}
		\makebox[\linewidth]{\resizebox{\linewidth}{!}{%
				\begin{tabular}{ccccc}
					\toprule
					Model & \text{full} & -$\mathbf{z}$ & -\text{joint} & -\text{augment} \\
					\toprule
					Accuracy & $\mathbf{62.86}_{\pm0.57}$ & $60.72_{\pm1.27}$ & $51.60_{\pm2.12}$ &  $57.20_{\pm0.86}$ \\
					\bottomrule
	\end{tabular}}}}
	\vspace*{-.5em}
	\caption{\label{tab:ablation_results} Accuracy on the CoNLL nominal devel data.}
	\vspace*{-.5cm}
\end{table}	
\subsubsection{Ablation study}\label{sec:supervised_ablation}

We investigate 
the importance of the latent code $\z$, joint modeling of arguments (\textit{joint}), and data augmentation (\textit{augment}).
We consider the VAE model with all the three components as the \textit{full} model.
Then we ablate the model by removing each of the three components individually.
We train each model five times with different random seeds 
and report the mean and unbiased standard deviation of model accuracies. 
Table~\ref{tab:ablation_results} summarizes the results. 
We can see that all the three components are indispensable. 
The smaller standard deviation of the \textit{full} model also suggests that they help stabilize the training process.  

\begin{figure}
	\begin{center}
		\includegraphics[width=1.\linewidth]{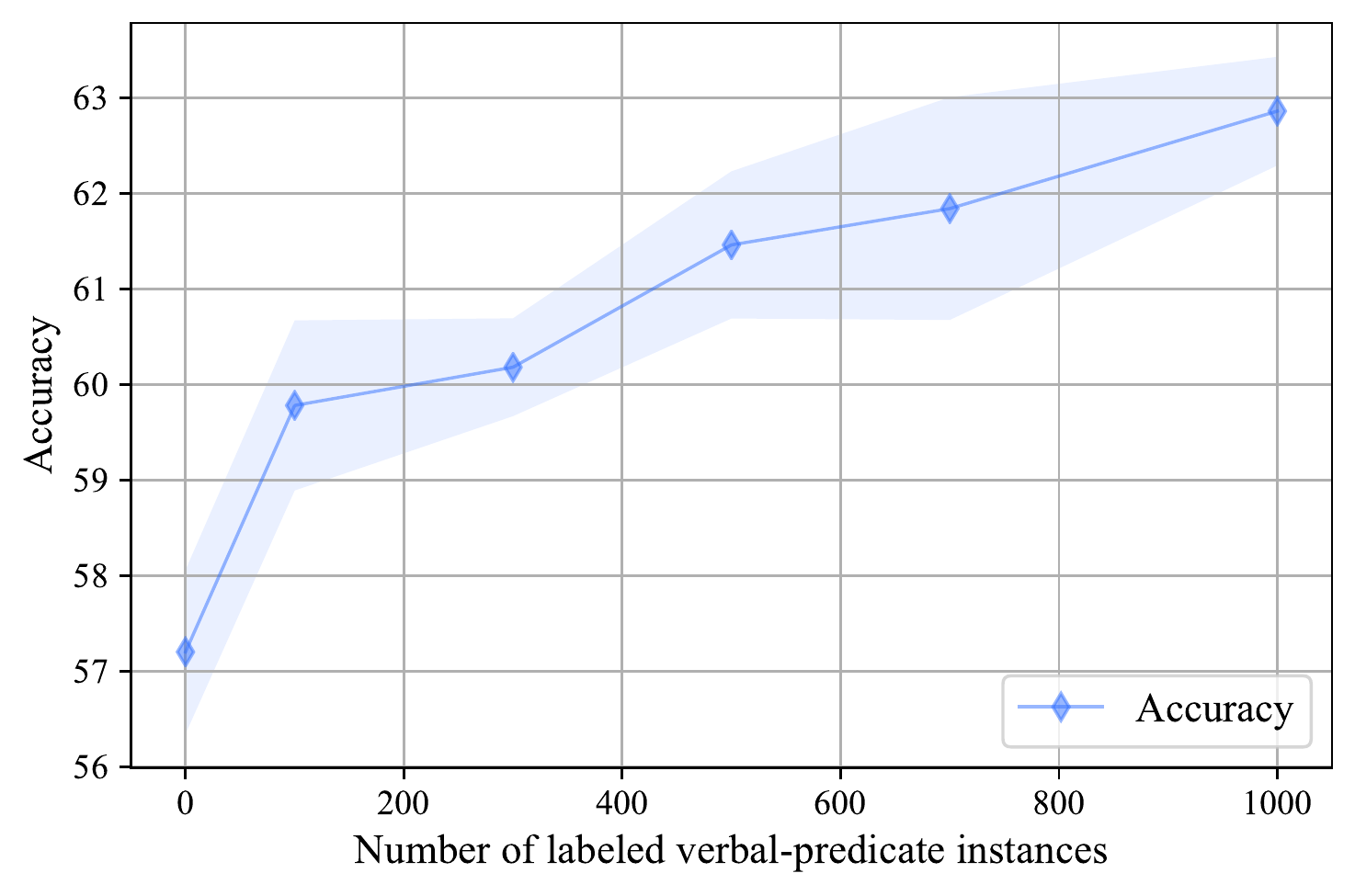}
	\end{center}
	\vspace*{-1em}
	\caption{\label{fig:data_size} Accuracy on the CoNLL nominal devel data.}
	\vspace*{-.5cm}
\end{figure}

We also study the impact of augmented-data size on the model performance.
Figure~\ref{fig:data_size} illustrates that the model accuracy consistently grows when increasing the number of labeled verbal examples (per predicate type).
We leave using more data for augmentation for future work.

\begin{table}[!t]\small
	\centering
	\vskip .0in
	\renewcommand{\arraystretch}{1.0}
	{\setlength{\tabcolsep}{1.em}
		\makebox[\linewidth]{\resizebox{\linewidth}{!}{%
				\begin{tabular}{rrrrr}
					\toprule
					Models & PU & CO & F1 & \\
					\toprule
					\multirow{5}{*}{\rotatebox[origin=c]{90}{}} 
					\text{AllA0} & 45.76 & 100.0 & 62.79  \\
					\text{SyntFun} & 63.42 & 66.81 & 65.07  \\
					\text{Arg2vec} & 64.31 & 64.36 & 64.33 \\
					\text{Factorization} & 75.29 & 52.77 & 62.05 \\
					\text{Most-frequent} & 66.48 & 75.10 & 70.53  \\
					\text{Direct-transfer} & 66.43 & 69.33 & 67.85  \\
					\toprule
					\text{Ours} & 71.38 & 78.56 & \textbf{74.80} & \\
					\bottomrule
	\end{tabular}}}}
	\vspace*{-.5em}
	\caption{\label{tab:unsupervised_results} Clustering results on the CoNLL nominal training data.}
	\vspace*{-.25cm}
\end{table}

\subsection{Unsupervised evaluation}\label{sec:unsupervised_eval}


\subsubsection{Baseline models and evaluation metrics}
We consider the following baseline models:
\begin{itemize}[wide=0\parindent,noitemsep]
	\item[1.]\text{\it AllA0} assigns all arguments to A0,
	the most frequent role in the data.
	\item[2.]\text{\it SyntFun} is a standard baseline in verbal semantic role induction.
	Each cluster corresponds to a syntactic function (i.e., a dependency relation 
	of an argument to its syntactic head).
	\item[3.]\text{\it Arg2vec} exploits the similarities among arguments in the embeddings space.
	It uses the agglomerative algorithm to cluster the pre-trained lemma embeddings of arguments.
	To be comparable with our model,
	we set the cluster number to 15, the number of roles in the verbal data. 
	\item[4.]\text{\it Factorization} is the verbal semantic role induction method of ~\citet{titov2015unsupervised}, which achieves state-of-the-art performance on the verbal domain. We use their code.
	\item[5.]\text{\it Most-frequent} and \text{\it Direct-transfer} are the supervised models from Section~\ref{sec:supervised_baselines} but evaluated using  unsupervised metrics.
\end{itemize}

\noindent\textbf{Evaluation metrics}: 
As standard in role induction literature, 
we evaluate models using clustering metrics:  purity (PU), collocation (CO), and their harmonic mean F1~\cite{lang2010unsupervised}.

\subsubsection{Experimental results}
Table~\ref{tab:unsupervised_results} presents clustering results of different models.
We can see that \textit{SyntFun} surpasses \textit{AllA0} by 2.28 in F1 measure,
suggesting that syntactic functions remain good predictors of semantic roles in the nominal domain.
However, \textit{Factorization} does not work well when applied to the nominal data.
Our model performs best,
ourperforming the strongest baseline, \textit{Most-frequent}, by 4.27\% F1.

\section{Additional related work}

\noindent\textbf{Transfer learning}, while popular for SRL, has focused on cross-lingual transfer.
Work in this direction 
uses annotation projection~\citep{pado2009cross,van2011scaling,aminian2019cross} 
or  relies on shared feature representation~\citep{kozhevnikov-titov-2013-cross}.
We are the first, as far as we know, to consider transfer across linguistic constructions.

\noindent\textbf{Semi-supervised Learning with VAEs}
is proposed by~\citet{kingma2014semi} for classification problem.
The VAEs have been adapted for various tasks in semi-supervised setting.
Most of them consist of both continuous and discrete latent variables~\citep{xu2017variational,zhou2017multi,chen-etal-2018-variational}.
Our model follows this line of work.
However, we focus on a transfer learning task 
since the labeled and unlabeled data come from different domains.


\section{Conclusion}

We defined a new task of transferring semantic role models 
from the labeled verbal domain to the unlabeled nominal domain.
We proposed to exploit the similarity in selectional preferences
across the domains and introduced a VAE model realizing this intuition. 
Our model outperformed both  supervised and unsupervised baselines.

In the future, we would like to explore alternative approaches
to transfer learning (e.g., using paraphrase models or pivoting via parallel data), and consider
other languages and other linguistic constructions (e.g., prepositions~\cite{srikumar2013modeling}).
It would also be interesting to use SemLink~\cite{palmer2009semlink} or AMR~\cite{banarescu2013abstract}  to evaluate transfer quality.

\section*{Acknowledgments}
We would like to thank Diego Marcheggiani for initial discussion which sets up the task proposed in this work.
We also want to thank Caio Corro, Nicola De Cao, Xinchi Chen, and Chunchuan Lyu for their helpful suggestions. The project was supported by the
European Research Council (ERC Starting Grant BroadSem
678254) and the Dutch National Science Foundation
(NWO VIDI 639.022.518).

\bibliographystyle{acl_natbib}
\bibliography{srl}

\section*{\centering Supplementary Material}

\begin{abstract}
	This supplementary material includes 
	(1) Reparameterization tricks for estimating the evidence lower bounds;
	(2) Processing details of the CoNLL data and the NYT data;
	(3) Configurations of the VAE model and the supervised \textit{Factorization} baseline model;
	and (4) Heuristics for identifying arugments of nominal predicates.
\end{abstract}

\appendix

\section{Reparameterization trick}\label{app:supp_learning}

We use gradient-based optimization to maximize the evidence lower bounds.
It involves sampling from  $q_{\phi}(\mb{y} | \mb{w})$ and $q_{\phi}(\mb{z} | \mb{w})$,
which is not differentiable,
This, together with the presence of discrete latent variables makes the optimization challenging.
As a remedy, 
for $\mb{z}$ we employ the reparametrization trick~\citep{kingma2014semi}
and define $\mb{z}$ as a deterministic variable:
\begin{flalign}
\mb{z} = \mb{\mu}_{\phi}(\mb{w}) + \sqrt{\mb{\sigma}_{\phi}(\mb{w})} * \mb{\epsilon}, \quad \mb{\epsilon} \sim \mathcal{N}(\mb{0}, \mb{I})\,.\nonumber
\end{flalign}
This sampling process is equivalent to sampling $\mb{z}$ from $\mathcal{N}(\mb{z} | \mb{\mu}_{\phi}(\mb{w}), \text{diag}(\mb{\sigma}_{\phi}(\mb{w})))$ directly.
However, it removes the non-differentiable sampling process from gradient back-propagation path.
For $\mb{y}$ we resort to the Gumbel-Softmax relaxation~\citep{jang2016categorical,maddison2016concrete},
which relies on the Gumbel-Max~\cite{maddison2014sampling} to draw samples $y$ from a categorical distribution over the role label space:
\begin{flalign}
y = \argmax_i [g_i + \log\pi_i],\quad g_i \sim \text{Gumbel}(0, 1)\,,\nonumber
\end{flalign}
where for $i\in\{1,\ldots,n_r\}$, $\pi_i$ specifies the probability $i$-th role label and $\sum_i\pi_i=1$; $g_i$ is drawn independently from the Gumbel distribution.
To avert the non-differentiable $\argmax$,
we represent $y$ as one-hot vector and 
approximate it using the softmax function: 
$y_i = \frac{1}{Z}\exp(g_i + \log\pi_i)$
where $Z = \sum_i \exp(g_i + \log\pi_i)$.

\begin{figure}[!t]
	\begin{center}
		\includegraphics[width=1.0\linewidth]{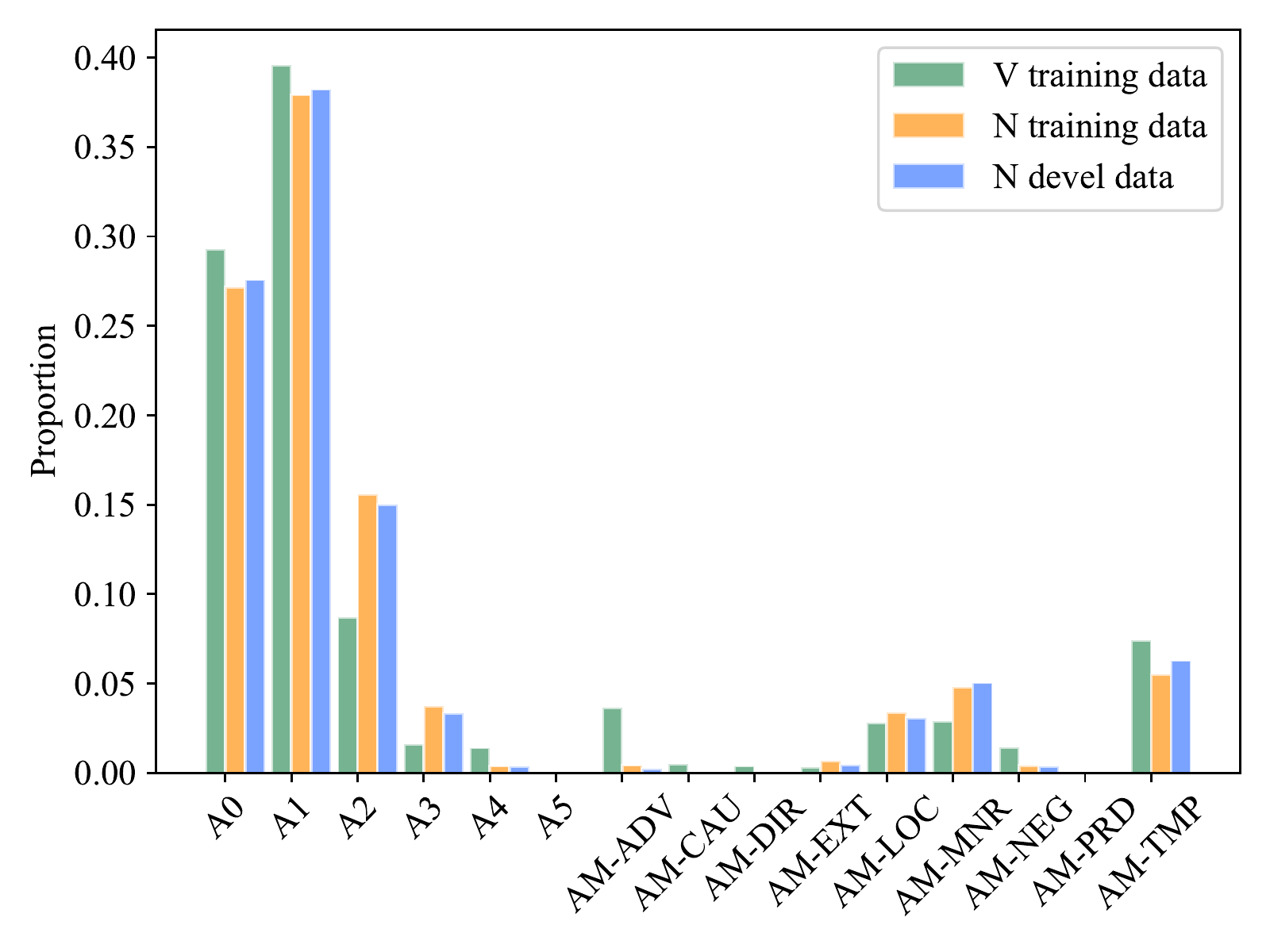}
	\end{center}
	\vspace*{-.75em}
	\caption{\label{fig:freq_roles}Proportion of each role in the verbal (V) training data and the nominal (N) data.}
\end{figure}

\section{Data}

\subsection{CoNLL data processing}\label{app:data_conll}
We apply verbalization to nominal predicates of the English CoNLL-2009 (CoNLL) dataset~\citep{hajivc2009conll}, mapping nouns to their verb forms according to the morph-verbalization list
\footnote{\url{http://amr.isi.edu/download/lists/verbalization-list-v1.06.txt}} .
Then we select a set of shared verb predicates between the two domains
and retain only instances having these shared predicates.
Moreover, we keep only the following 15 frequent role labels in the two domains (a cut-off of 3): 
A0, A1, A2, A3, A4, A5, AM-ADV, AM-CAU, AM-DIR, AM-EXT, AM-LOC, AM-MNR, AM-NEG, AM-PRD, AM-TMP.
Statistics of the resulting datasets are shown in Table~\ref{tab:conll_tf_data_stats}.
Figure~\ref{fig:freq_roles} shows proportions of the 15 roles in the final data.
In training,
we use only the predicates attested with at least 20 instances in the nominal domain and at least 10 instances in the verbal domains,
This rules out infrequent predicates which either trivially lead to a high score or lack labeled data to learn from.


\begin{table}[!t]
	\centering
	\vskip .0in
	{\setlength{\tabcolsep}{.25em}
		\makebox[\linewidth]{\resizebox{\linewidth}{!}{%
				\begin{tabular}{crrrrrr}
					\toprule
					\multicolumn{1}{c}{} &
					\multicolumn{2}{c}{train} &
					\multicolumn{2}{c}{dev} &
					\multicolumn{2}{c}{test} \\
					\toprule
					& \# pred-  &  \# sample & \# pred-  &  \# sample & \# pred-  &  \# sample\\
					\toprule
					V   & 1025 & 68461 & 387 & 1404 & 454 & 2223 \\
					N   & 1025 & 38283  & 533 & 2456 & 582 & 3955 \\
					\bottomrule
	\end{tabular}}}}
	\caption{\label{tab:conll_tf_data_stats} Number of instances (\# sample) and predicate forms (\# pred-) in the verbal (V) and nominal (N) domains of the processed CoNLL data.}
\end{table}

\subsection{Verbal data augmentation}\label{app:data_augmentation}
We pre-train a semantic role labeler on the CoNLL verbal training data.
The model gives rise to an F1 score of 89.2 on the CoNLL verbal development data, which is strong enough.
Then we apply it to the verbal predicates in the New York Times Annotated Corpus (NYT)~\citep{sandhaus2008new} and obtain extra labeled verbal data.

The preprocessing of the NYT data is as follows:
first we tokenize and segment the NYT data into sentences and rule out sentences having more than 45 words.
Then we lemmatize each sentence and label it with part-of-speech (POS) tags using a pre-trained lemmatizer~\citep{muller2015joint}.
After that, we employ the following heuristic to identify predicates in the sentences:
we collect all pairs of predicate and its POS tag from the CoNLL training data,
then we search for matches to these pairs of predicate and POS tag in the NYT data.
For every complete match with the same predicate and POS tag, we count it as a predicate occurrence.
Finally, we extract verbal-predicate data and retain instances which share the same set of verbal predicates as in the processed CoNLL data.

\section{Model parameters}\label{app:model_params}

The semantic role model $q_{\phi}(\mb{y}|\cdot)$ uses the same highway BiLSTM encoder as in~\citet{he2017deep}.
The encoder model $q_{\phi}(\mb{z}|\cdot)$ also uses a highway BiLSTM but with only one layer (interested readers are referred to~\citet{he2017deep}).

\subsection{Supervised baseline model}\label{app:sup_bl}
\noindent\textbf{\it Factorization} estimates the compatibility of a role $y_i$ with predicate $p$ and argument $a_i$:
\begin{flalign*}
s(a_i, y_i, p) = \mb{v}_{a_i}^{T}\mb{W}_{y_i}\mb{v}_{p} + \mb{v}_{a_i}^{T}\mb{w}_{y_i}\,,
\end{flalign*} 
where $\mb{v}_{a_i}$ and $\mb{v}_{p}$ are argument and predicate embeddings, respectively;
$\mb{W}_{y_i}$ and $\mb{w}_{y_i}$ is a matrix embedding and a vector embedding of the role, respectively.
The model is trained to maximize the conditional log-likelihood of $a_i$ given $p$ and $y_i$:
\begin{flalign*}
\log p(a_i | p, y_i) = \log \frac{\exp(s(a_i, y_i, p))}{\sum_{j}\exp(s(a_j, y_i, p))}\,.
\end{flalign*}
We use the pre-trained lemma embeddings for arguments and randomly initialize predicate embeddings.
Both the embeddings are of dimension 100 and are tunable during training.

\section{Argument identification}\label{app:heuristics}

We use the following five rules to heuristically identify arguments for nominal predicates. 
The heuristics rely only on gold syntactic dependency structures and part-of-speech tags. 
Initially, all words in a sentence are treated as non-arguments.

\begin{itemize}[wide=0\parindent,noitemsep]
	\item[1.] Determinators and punctuations are removed from consideration
	\item[2.] Any of a predicate's dependents that appears before the predicate are labeled as  arguments (i.e., the dependent's word index is smaller than the predicate's word index). 
	When it appears after the predicate, it is labeled as an argument if and only if it is a preposition.
	\item[3.] All preposition siblings of a predicate are labeled as arguments if and only if the predicate does have any dependents.
	\item[4.] All SBJ siblings of a predicate are labeled as arguments if and only if the predicate's relation to its head is OBJ.
	\item[5.] The closest word to a predicate that appears before the predicate and has the dependency label SBJ is labeled as an argument if and only if the predicate does not have any dependents preceding the predicate.
\end{itemize}

\end{document}